\documentclass[10pt,twocolumn,letterpaper]{article}

\usepackage{cvpr}
\usepackage{times}
\usepackage{epsfig}
\usepackage{graphicx}
\usepackage{amsmath}
\usepackage{amssymb}
\usepackage{float}
\cvprfinalcopy % *** Uncomment this line for the final submission

\def\cvprPaperID{****} % *** Enter the CVPR Paper ID here

\begin{document}

%%%%%%%%% TITLE

\title{GAN-Knowledge Distillation for one-stage Object Detection}

\author{Wei Hong\\
Startdt AI Lab\\
{\tt\small 517332051@qq.com}
\and
Jinke Yu\\
Zeusee Technology\\
{\tt\small jackyu@zeusee.com}
\and
Zong Fan\\
Startdt AI Lab\\
{\tt\small fanzong@startdt.com}
}

\maketitle
%\thispagestyle{empty}
%%%%%%%%% ABSTRACT
\begin{abstract}
   Convolutional neural networks have a significant improvement in the accuracy of object detection. As convolutional neural networks become deeper, the accuracy of detection is also obviously improved, and more floating-point calculations are needed. Many researchers use the knowledge distillation method to improve the accuracy of student networks by transferring knowledge from a deeper and larger teachers network to a small student network, in object detection. Most methods of knowledge distillation need to designed complex cost functions and they are aimed at the two-stage object detection algorithm. This paper proposes a clean and effective knowledge distillation method for the one-stage object detection. The feature maps generated by teacher network and student network are used as true samples and fake samples respectively,and generate adversarial training for both to improve the performance of the student network in one-stage object detection.
.
\end{abstract}

%%%%%%%%% BODY TEXT
\section{Introduction}

In recent years, with the development of deep learning, researchers have found that using deeper and larger convolution neural network as the backbone of object detection, the accuracy of object detection is improved more. with the improvement of detection accuracy of object detection, computer vision moves from non-critical areas to key areas (such as unmanned driving and medical fields). However, in order to ensure the detection accuracy, a larger convolution neural network has to be used as backbone of object detection, resulting in a decrease in the detection speed and an increase in the cost of computing equipment. Therefore, many researchers propose many methods to improve the detection speed on the premise of ensuring the detection accuracy, such as the method of reducing the number of floating-point operations of convolution neural network by depth-wise separable convolution~\cite{MobileNets,MobileNetV2}point wise group convolution and channel shuffle~\cite{shufflenetv1,shufflenetv2}. Although considerable speed-up results have been achieved, these methods require carefully designed and tuned backbone of network.
Many researchers believe that although the deeper backbone of network has a larger network capacity, so in the image classification, object detection and other tasks have a better performance. However, some specific tasks do not need such a large capacity, so in the case of ensuring the accuracy of convolution neural network, compression, quantization, channel subtraction on convolutional neural networks~\cite{5,6,7,8,9}.

On the other hand, there are some studies on knowledge distillation that show~\cite{10, 11, 12, 13}, use a deeper model and a lighter model as teacher net and student net, then training student net with true label and soft label that teacher net output or intermediate result of teacher net output. It can greatly improve the performance of student net on specific tasks. But most of these methods require very complex cost functions and training methods, and these methods are mostly used for image classification, and two-stage object detection, etc., and are rarely used for one-stage object detection. Therefore, we need a knowledge distillation method that is simpler and more effective and can be applied to one-stage object detection. This paper proposes a simple and effective knowledge distillation neural network architecture, and it can obviously improve the performance of student net in one-stage object detection. Different from the conventional knowledge distillation method, We separated the backbone of deeper object detection neural network and lighter object detection neural network as teacher net and student net respectively, then we use the feature map generated by the teacher net as true samples, and the feature map generated by the student net as fake samples,referring to architecture of generative adversarial Nets~\cite{14}. Finally, we design a neural network as a discriminator and use true samples and fake samples to do the generation adversarial training.

\noindent
There are two main contributions:

1. Propose a clean and effective architecture of knowledge distillation that does not require the design of complex cost functions, and can be applied to one-stage object detection.

2. Using the architecture of generation adversarial nets to avoid complex knowledge migration design, let the student net automatically obtain dark knowledge from the teacher net.

\section{Related Works}
\noindent
{\bf CNN For Detection}: The deep learning architecture of object detection is mainly divided into two types:1) one is the one-stage object detection, such as the SSD proposed by Liu W et al. ~\cite{15}, which directly returns the position and category of the object by the convolutional neural network; 2)two is the two-stage object detection, such as fast rcnn ~\cite{16} proposed by Girshick et al., and later Faster-RCNN ~\cite{17} and R-FCN ~\cite{18}, etc., it first regress the proposal boxes by the convolutional neural network; then identify each proposal box again; Finally return to the correct location and category.

\noindent
{\bf Network Compression}: many researchers believe that deep neural networks are over-parameterized, and it has too many redundant neurons and connections. He Y et al. [8] thought each layer of neurons of convolutional neural networks are sparse, and they use lasso regression to find the most representative neuron per layer of convolutional neural networks reconstructs the output of this layer. Zhuang Z et al. ~\cite{9} believe that layer-by-layer channel pruning affects the discriminating ability of Convolutional neural networks, so the auxiliary ability of convolutional neural networks is preserved by adding auxiliary loss in the fine-tune and pruning stages.

\noindent
{\bf Network quantification}: Wu J et al. ~\cite{20} use the k-means clustering algorithm to accelerate and compress the convolutional layer and the fully connected layer of the model to obtain better quantization results by reducing the estimation error of the output response of each layer, and proposed an effective training scheme to suppress the multi-layer cumulative error after quantization. Jacob B ~\cite{21} et al. proposed a method that quantify weights and inputs as uint8, and bias to unit32, as the same time, the forward uses quantization, and the backward correction error is not quantized to ensure that the convolutional neural networks performance and speed of inference during training.

\noindent
{\bf Knowledge Distillation}: A method for transfer knowledge of large network to small network. Hinton et al. ~\cite{6} used the result of teacher net output as the soft label of student net, and advocated the use of temperature cross entropy instead of L2 loss.Romero et al. ~\cite{19} believe that student net needs more unlabeled data to get as close as possible and imitate teacher net.When Chen G et al. ~\cite{12} distilled the two-stage object detection, they respectively extracted the middle feature map of teacher net and the dark knowledge of rpn/rcnn to let student net go to mimic.There are also some researchers who give the attention information of teacher net to the student network. For example. For example, Zagoruyko S ~\cite{22} et al. proposed spatial-attention, which is a method of transmitting the thermal information of teacher net to student net. Yim J et al. ~\cite{23} used the relationship between layer and layer of teach net as the goal of student network learning. However their knowledge distillation requires the design of very complex loss functions, and the extraction of complex dark knowledge,and these methods are mostly in the two-stage object detection, rarely used in one-stage object detection. In order to have a simple and effective way of distilling knowledge, we refer to the architecture of the generated adversarial nets ~\cite{14} to take the feature map generated by the teacher network and the student network as true samples and fake samples, respectively. Finally we design a neural network as a discriminator and use true samples and fake samples to do the generation adversarial training to improve the performance of the student network in one-stage object detection.

\begin{figure*}
\begin{center}
\includegraphics[width=16cm, height =8cm]{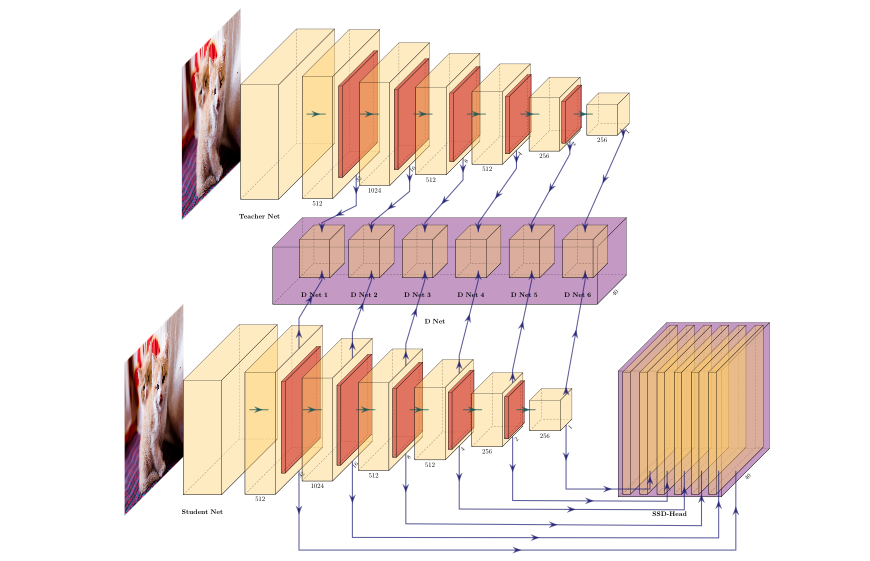}
\end{center}
  \caption{Teacher Net and SSD-Head are the backbone network and head network of the larger and fully trained SSD model, respectively. Student Net is a smaller network. D Net is a module consisting of multiple 6 small discriminant networks.}
\label{fig:short}
\end{figure*}

In this paper, we use the one-stage object detection SSD ~\cite{15} as our object detection. The architecture of SSD is mainly divided into two parts, 1) backbone of network, as feature extractor. 2) SSD-Head, use the features extracted by the backbone of network to detect the category and location of the object. In order to obtain a better knowledge distillation effect, it is important to make rational use of these two parts.

\section{Method}

\subsection{Overall Structure}
Fig 1 is the overall structure of our model. We first use a SSD model with a larger capacity and full training, and split the SSD model into backbone of network and SSD-Head. We use the backbone of network as the teacher net,and pick a smaller network as student net. We use multiple feature maps generated by teacher net as true samples, and multiple feature maps generated by student net as fake samples, then send the true sample and the fake sample to each of the corresponding discriminative networks (fig 2) in D Net,at the same time, input the fake sample into the SSD-Head.

\begin{figure}[H]
\begin{center}
\includegraphics[width=6cm, height =4cm]{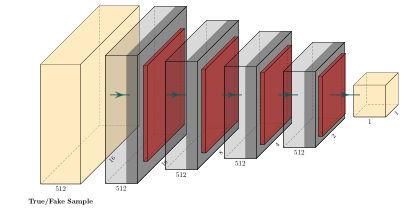}
\end{center}
  \caption{One of the discriminative networks in the D-Net module, consisting of multiple downsampled convolutional layers.}
\label{fig:short}
\end{figure}

\subsection{Training Process}
Our training process has two stages.The first stage is mainly the use of the generation adversarial training (GAN). First,we train each discriminating network in the D Net module to identify whether the input sample is true sample or fake sample, and freeze the weights of teacher net and student net. Then train the student net to generate a fake sample to trick each of the discriminating networks in the D Net module, and normal SSD training is performed on student net and SSD-Head.During the training process,weight of the teacher net and weights of each discriminative network in the D Net module are frozen,as Equation 1.
\begin{equation}\label{eq:12}
\begin{split}
& L\mathop{{}}\nolimits_{{Dt}}=D{ \left( {Teacher{ \left( {x, \theta \mathop{{}}\nolimits_{{t}}} \right) }, \theta \mathop{{}}\nolimits_{{d}}} \right) } \\
& L\mathop{{}}\nolimits_{{Ds}}=D{ \left( {Student{ \left( {x, \theta \mathop{{}}\nolimits_{{s}}} \right) }, \theta \mathop{{}}\nolimits_{{d}}} \right) } \\
& L\mathop{{}}\nolimits_{{conf}}=L\mathop{{}}\nolimits_{{conf}}{ \left( {Student{ \left( {x, \theta \mathop{{}}\nolimits_{{s}}} \right) }}, \theta \mathop{{}}\nolimits_{{head}} \right) } \\
& L\mathop{{}}\nolimits_{{loc}}=L\mathop{{}}\nolimits_{{loc}}{ \left( {Student{ \left( {x, \theta \mathop{{}}\nolimits_{{s}}} \right) }},\theta \mathop{{}}\nolimits_{{head}} \right) } \\
& L\mathop{{}}\nolimits_{{D}}=\frac{{1}}{{N}}{ \left( {L\mathop{{}}\nolimits_{{Dt}}-L\mathop{{}}\nolimits_{{Ds}}} \right) } \\
& L\mathop{{}}\nolimits_{{G}}=\frac{{1}}{{N}}{ \left( {D{ \left( {Student{ \left( {x, \theta \mathop{{}}\nolimits_{{s}}} \right) }, \theta \mathop{{}}\nolimits_{{d}}} \right) }+L\mathop{{}}\nolimits_{{conf}}+L\mathop{{}}\nolimits_{{loc}}} \right) } \\
\end{split}
\end{equation}

N in Equation 1 represents the size of batchsizes; D represents the discriminant network, Teacher and Student represent teacher net and student net, respectively; θt, θs, and θd represent the weight of the teacher net, the weight of the student net, and the weights of each discriminant network in the D Net module, respectively. Lconf represents the loss function of the classification in the SSD, and Lloc represents the loss function of the bounding box in the SSD. The second stage is that we normal SSD training on student net and SSD-Head alone,after many epochs of generation adversarial training .

%------------------------------------------------------------------------
\section{Experiment}
In this section, we will experiment with the PASCAL VOC and cocoto validate our approach, including 20 categories.  Our hardware device is two NVIDIA GTX 1080Ti GPUs. The software framework is gluoncv.
\subsection{Training Process}
All models used adam with 0.0005 learning rate in the first phase and SGD with 0.0005 weight decay and 0.9 momentum in the second phase. The first stage trains epochs is 180 and the second stage epochs is 90. The stduent net is MobilenetV1, MobilenetV2 and resnet18, teacher net VGG16, ResNet50 and ResNet101. These models have  pretrained under ImageNet. Due to insufficient time, we only use MobilenetV1 as Student Net and ResNet50 as Teacher Net do experiment on coco. 
\subsection{Results}
We compare the native SSD with the SSD of GAN knowledge distillation under different Teacher nets,up to the student net 2.8mAP on PASCAL VOC. When the teacher net is ResNet101 and student net is ResNet18, the improvement  is not as good as teacher net is ResNet50.Because precision of SSD of ResNet101  is not as good as  precision of SSD of ResNet50 on PASCAL VOC,and SSD of ResNet101 is only one mAP higher than SSD of ResNet50 on coco.So in order to complete the experiment efficiently,We only use ResNet50 as teacher net on coco,and we will use different teacher net in the near future.

From the table.1, we found that the lower the mAP on the data set, the more obvious the effect of the student net after GAN knowledge distillation.So we use moblienetV0.75 as the student net on coco,and We have increased 5 mAP for SSD of moblienetV0.75 on coco.
\begin{table}
\begin{center}
\begin{tabular}{|l|l|c|}
\hline
Student net & Teacher net & voc 2007 test\\
\hline\hline
            & - & 75.4 \\
MobilenetV1 & VGG16 & 77.3(+1.9)\\
            & ResNet50 & 77.6(+2.2)\\
            & ResNet101 & 77.6(+2.2)\\
\hline\hline
            & -         & 75.9   \\
MobilenetV2 & VGG16     & 77.2(+1.4)\\
            & ResNet50  & 77.7(+1.7)\\
            & ResNet101 & 77.5(+1.5)\\
\hline\hline
            & - & 74.8 \\
ResNet18    & VGG16 & 77.2(+2.6)\\
            & ResNet50 & 77.6(+2.8)\\
            & ResNet101 & 77.3(+2.7)\\
\hline
\end{tabular}
\end{center}
\caption{Different student nets are not used GAN-knowledge distillation and the use of a GAN-knowledge }
\end{table}
\begin{table}
\begin{center}
\begin{tabular}{|l|l|c|}
\hline
Student net & Teacher net & coco 2017 val\\
\hline\hline
            & - & 21.7 \\
MobilenetV1 & VGG16 & -\\
            & ResNet50 & 25.4(+3.7)\\
            & ResNet101 & - \\
\hline\hline
                                          & - & 18.0 \\
MobilenetV1\underline{\hspace{0.5em}}0.75 & VGG16 & -\\
                                          & ResNet50 & 23.0(+5.0)\\
                                          & ResNet101 & - \\

\hline\hline
            & -         & 22   \\
MobilenetV2 & VGG16     & -\\
            & ResNet50  & 24.6(+2.6)\\
            & ResNet101 & -\\
\hline
\end{tabular}
\end{center}
\caption{ moblienetv1 use GAN-knowledge distillation in coco.}
\end{table}

We also use our method to improve the two stage object detection,such as faster rcnn.We found that faster rcnn of roialign is 4.7 mAP higher than  faster rcnn of roipooling in Pascal VOC 2007 test,as shown in Table 3.Our method gets 6.8 absolute gain in mAP for faster rcnn of ResNet50.

% faster rcnn table
\begin{table}
\begin{center}
\begin{tabular}{|l|l|c|}
\hline
Student net & Teacher net & voc 2007 test\\
\hline\hline
            & - & 67.0 \\
ResNet50(roipooling+GAN-KD) & ResNet101 & 73.8(+6.8)\\

\hline\hline
            & - & 71.7 \\
ResNet50(roialign+GAN-KD) & ResNet101 & 74(+2.3)\\
\hline\hline
            & - & 69.0 \\
ResNet50(Imitation) & ResNet101 & 72.0(+3.0)\\
\hline
\end{tabular}
\end{center}
\caption{ Teacher net is the faster rcnn of the ResNet101, and uses Pascal Voc 2007 trainval as the training set. The mAP is 74.8+ on the Pascal Voc 2007 test set. The first row and the second row is our method, and the third row is Distilling Object Detectors with Fine-grained Feature Imitation\cite{wang2019distilling}}
\end{table}
%------------------------------------------------------------------------
\section{Conclusion}
At present, most of the methods of knowledge distillation are aimed at the two-stage object detection. We propose a clean and effective knowledge distillation method for the one-stage object detection. The feature map generated by the teacher net is taken as  true samples, and the feature map generated by the student net is taken as fake samples. Then We make the distribution of student net to fit the distribution of teacher net by using true samples and fake samples to do the generation adversarial training. Through unsupervised learning then generation adversarial training, we avoid manual design and extract feature from teacher net to let the student net fit. In this way, we make the whole training process simple and efficient, and greatly improve the detection accuracy of the student net. Our approach provides new ideas for the further development of knowledge distillation.

{\small
% \bibliographystyle{ieee_fullname}
% \bibliography{egbib}

}

\end{document}